\documentclass[10pt,twocolumn]{article}
\usepackage[margin=2cm,columnsep=0.8cm]{geometry}
\usepackage{graphicx}
\usepackage{amsmath,amssymb}
\usepackage{booktabs}
\usepackage{multirow}
\usepackage[hidelinks]{hyperref}
\usepackage{xcolor}
\usepackage[numbers,sort&compress]{natbib}
\usepackage{microtype}
\usepackage{caption}
\usepackage{stfloats}
\usepackage{algorithm}
\usepackage{algpseudocode}
\usepackage{titlesec}
\titlespacing*{\section}{0pt}{1.6em}{0.8em}
\titlespacing*{\subsection}{0pt}{1.2em}{0.6em}
\titlespacing*{\paragraph}{0pt}{0.9em}{0.9em}

\newcommand{\snq}{S\&Q}
\newcommand{\fyes}{\checkmark}
\newcommand{\fno}{\ensuremath{\times}}
\newcommand{\fpart}{(\checkmark)}

\begin{document}

\twocolumn[{%
\begin{center}
{\LARGE\bfseries Signals and Queries:\\[0.25em] Token-Efficient Agentic Context Discovery\par}
\vspace{1.1em}
{\large Yam Marcovitz}\\[0.25em]
\texttt{yam@parlant.io}
\vspace{2.4em}

{\normalsize\bfseries Abstract\par}
\vspace{0.7em}

\begin{minipage}{0.86\textwidth}
\fontsize{10}{13}\selectfont
User-facing LLM agents must identify which behavioral policies apply at each turn, but evaluating every policy with an LLM is costly. We introduce Signals and Queries (\snq{}), a lightweight first-stage relevance filter developed for Parlant~4.0 that removes at least 30\% of candidate policies before downstream evaluation while preserving high recall. For each policy, an LLM generates activating examples and hard negatives offline. An independent logistic classifier trained on their embeddings then predicts relevance using embeddings and vector dot products, with no per-turn LLM calls. This shifts expensive relevance evaluation to a smaller candidate set, while allowing each policy's classifier to be updated independently. Evaluated against 9{,}144 LLM-judge relevance labels covering 254 $\tau^2$-bench conversation prefixes and 36 airline policies, \snq{} achieves higher precision than top-$K$ and cosine-threshold retrieval at every evaluated recall level, with consistent results across two embedding providers. These findings support inexpensive, per-policy classification as a practical first stage for reducing downstream evaluation workload in customer-facing agents.
\end{minipage}
\vspace{3.0em}
\end{center}
}]

\section{Introduction and Motivation}

Agentic LLM applications increasingly separate \emph{what the agent could know} from \emph{what the agent should see right now}. A customer-facing agent may be governed by dozens of behavioral policies; it may also carry a domain glossary, tool and skill descriptions, state observations, and reference documents. Placing all of it in every prompt is expensive and counterproductive, for three well-documented reasons. Long contexts suffer from the \emph{lost-in-the-middle} effect: models systematically under-attend to information positioned in the middle of the prompt~\citep{liu2024lost}. They suffer from \emph{attention dilution}: as the prompt grows, the attention mass allotted to any individual instruction shrinks, so its influence on generation fades~\citep{tian2024spa}. And they suffer from the \emph{curse of instructions}: the probability of following all instructions decays approximately as the per-instruction success rate raised to the number of instructions, so adherence degrades steadily as inventories grow~\citep{harada2025curse}. On top of these quality failures, per-turn token cost scales linearly with inventory size rather than with what the turn actually needs.

The problem is acute for \emph{policies}. In behavioral governance frameworks such as Parlant~\citep{parlant}, policies are normative instructions (``only compensate if the customer holds travel insurance\ldots'') whose \emph{omission} is a correctness failure, not merely a quality regression: an agent that does not see the cancellation policy while handling a cancellation may take actions it should have refused. Relevance filtering for policies is therefore a \emph{recall-critical} problem: a false negative can produce a policy violation, while a false positive merely wastes a few hundred prompt tokens that modern models tolerate well. This asymmetry drives the design of everything that follows.

One accurate solution is to ask an LLM, per turn and per policy, ``is this policy relevant now?''; this is the first-pass \emph{ranker} used in Parlant's engine. It works, but its cost scales with the inventory: for our 36-policy benchmark it consumes ${\sim}69{,}000$ prompt tokens and 36 LLM calls \emph{per user turn} (\S\ref{sec:cost}). The alternative, embedding-based retrieval of the policy text against the conversation, is nearly free but suffers from a genre gap: conversations are informal user utterances while policies are formal normative prose, and their embeddings separate poorly~\citep{ethayarajh2019contextual}. A third architecture avoids the problem by construction: routed flow-graphs place different instructions at different nodes, so each prompt stays small. But this fragments the agent's knowledge. Real customer conversations are multi-turn and multi-topic, and drift freely across node boundaries. The node in control at any given moment often lacks policies the situation has come to require, so the agent is least coherent on the messy, multi-topic conversations that matter most. A fourth approach, taken by skill-based agent frameworks~\citep{anthropic2025skills,openai2026codexskills}, keeps only a short description of each skill resident and loads a skill's full instructions on demand, once the model, reasoning over those descriptions, decides it applies. This keeps whole instructions out of the prompt, but makes discovery part of the model's own reasoning loop, slower and costlier than an out-of-loop lookup.

We propose \emph{Signals and Queries} (\snq{}): describe each artifact not (only) by its text but by \emph{signals} (LLM-generated examples of user messages that should activate it) and \emph{anti-signals} (hard near-misses that should not). A tiny per-artifact classifier trained on these exemplars then scores live conversation turns (\emph{queries}) in the embedding space. While we develop and evaluate \snq{} on agent policies, nothing in the formulation is policy-specific: any context artifact that can be described by exemplar activating queries is a candidate target, including state observations, glossary entries, skills and tools~\citep{patil2023gorilla,qin2023toolllm}, and code or document sections.

\paragraph{Contributions.}
(1)~A formulation of agentic context discovery as recall-first, per-artifact binary relevance, decided by classifiers over LLM-generated signals and anti-signals; the design is \emph{decoupled}: each artifact's classifier and threshold depend only on its own exemplars, so adding or editing an artifact retrains nothing else.
(2)~A benchmark against an LLM judge on $\tau^2$-bench conversations~\citep{yao2024taubench,barres2025tau2} with 36 policies, comparing four embedding-retrieval baselines; \snq{} Pareto-dominates all of them and replicates across two embedding providers.
(3)~A cost analysis showing a ${\sim}10^5\times$ reduction in per-turn cost relative to per-policy LLM ranking.
(4)~An analysis of failure modes showing that signal and anti-signal \emph{quality} (coverage, hardness, and domain match) is the dominant factor, motivating exemplar generation as a research problem in its own right.

This work continues our ongoing pursuit of accuracy in customer-facing agents at Parlant. It complements our earlier work on Attentive Reasoning Queries~\citep{karov2025arq}, which optimizes instruction-following once the relevant policies are already in the agent's context; \snq{} addresses the upstream problem of discovering which policies belong there in the first place.

\section{The \snq{} Method}
\label{sec:method}

\subsection{Definitions}
Let $\mathcal{A}=\{a_1,\dots,a_N\}$ be the artifact inventory (policies). For each artifact $a_j$ an LLM generates, offline:
\begin{itemize}\itemsep2pt
  \item \textbf{Signals} $S_j = \{s_{j1},\dots,s_{jm}\}$: short, natural user messages that \emph{should} activate $a_j$ (``I need to cancel my flight to Chicago next week'').
  \item \textbf{Anti-signals} $\bar S_j = \{\bar s_{j1},\dots,\bar s_{jm}\}$: \emph{hard near-misses}, i.e., messages in the same topical neighborhood that should \emph{not} activate $a_j$ because a neighboring artifact governs them (for a cancellation policy: ``I need to change my flight from Tuesday to Thursday'').
\end{itemize}
We use $m=10$ of each. This is the document-side analogue of query expansion: signals play the role of predicted queries as in Doc2Query~\citep{nogueira2019doc2query}, and are generated without any conversation data, as in synthetic-query domain adaptation~\citep{wang2022gpl}; anti-signals additionally supply \emph{targeted hard negatives}, which cosine similarity alone cannot exploit.

A \textbf{query} is the live conversation at its current state. At turn $t$ we form two kinds of view, both centered on the artifact-inventory centroid $c$ (centering mitigates the anisotropy of embedding spaces~\citep{mu2018allbutthetop,ethayarajh2019contextual}): the \emph{cumulative} view $q_{\mathrm{cum}}^{(t)}$, the embedding of the whole conversation through turn $t$, and a \emph{user-turn} view $q_{\mathrm{user}}^{(u)}$, the embedding of the single user message sent at each turn $u \le t$. All embeddings are $\ell_2$-normalized.

\subsection{Per-artifact classifier}
\begin{figure*}[t]
  \centering
  \includegraphics[width=0.78\textwidth]{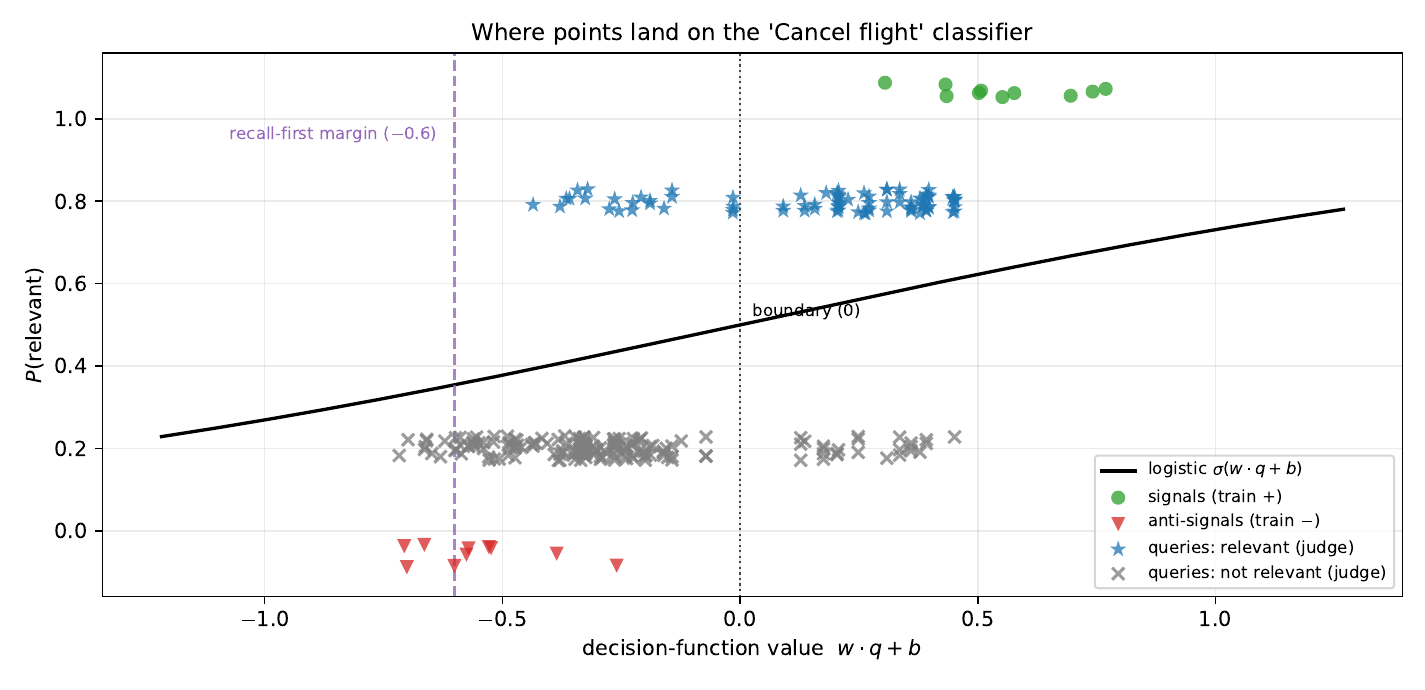}
  \caption{Where points land on one artifact's logistic classifier (\emph{Cancel flight}). Signals and anti-signals separate cleanly; real conversation queries labeled relevant by the LLM judge (stars) score above the irrelevant ones (crosses) but \emph{below} the signals. This signal$\to$query domain shift motivates the recall-first margin $\theta<0$.}
  \label{fig:logreg}
\end{figure*}

For each artifact independently, we fit a logistic regression on the embedded exemplars: positives $S_j$, negatives $\bar S_j$:
\begin{equation}
 P(\text{relevant}\mid q) = \sigma(w_j\!\cdot\!q + b_j).
\end{equation}
At inference, the score of artifact $a_j$ for the conversation at turn $t$ is a max-pool over query views,
\begin{equation}
\label{eq:maxturn}
 z_j(t) = \max\Big(w_j\!\cdot\!q_{\mathrm{cum}}^{(t)},\; \max_{u \le t} \, w_j\!\cdot\!q_{\mathrm{user}}^{(u)}\Big) + b_j,
\end{equation}
and $a_j$ is retrieved iff $z_j(t) > \theta$. The max over user turns keeps a policy discoverable from the turn that activates it, regardless of what the conversation discusses before or after: a single blended embedding of a long, multi-topic conversation dilutes any one topic, whereas the per-turn max preserves each turn's own evidence. On the short, single-topic conversations of our benchmark this makes no difference, but it matters when the conversation shifts topic, which we quantify in \S\ref{sec:ablation}. The margin $\theta$ is a single global knob trading recall for precision; we operate recall-first at $\theta<0$ because real conversation turns score systematically lower than clean, idealized signals (a visible domain shift; Fig.~\ref{fig:logreg}).

\begin{figure*}[t]
  \centering
  \includegraphics[width=0.9\textwidth]{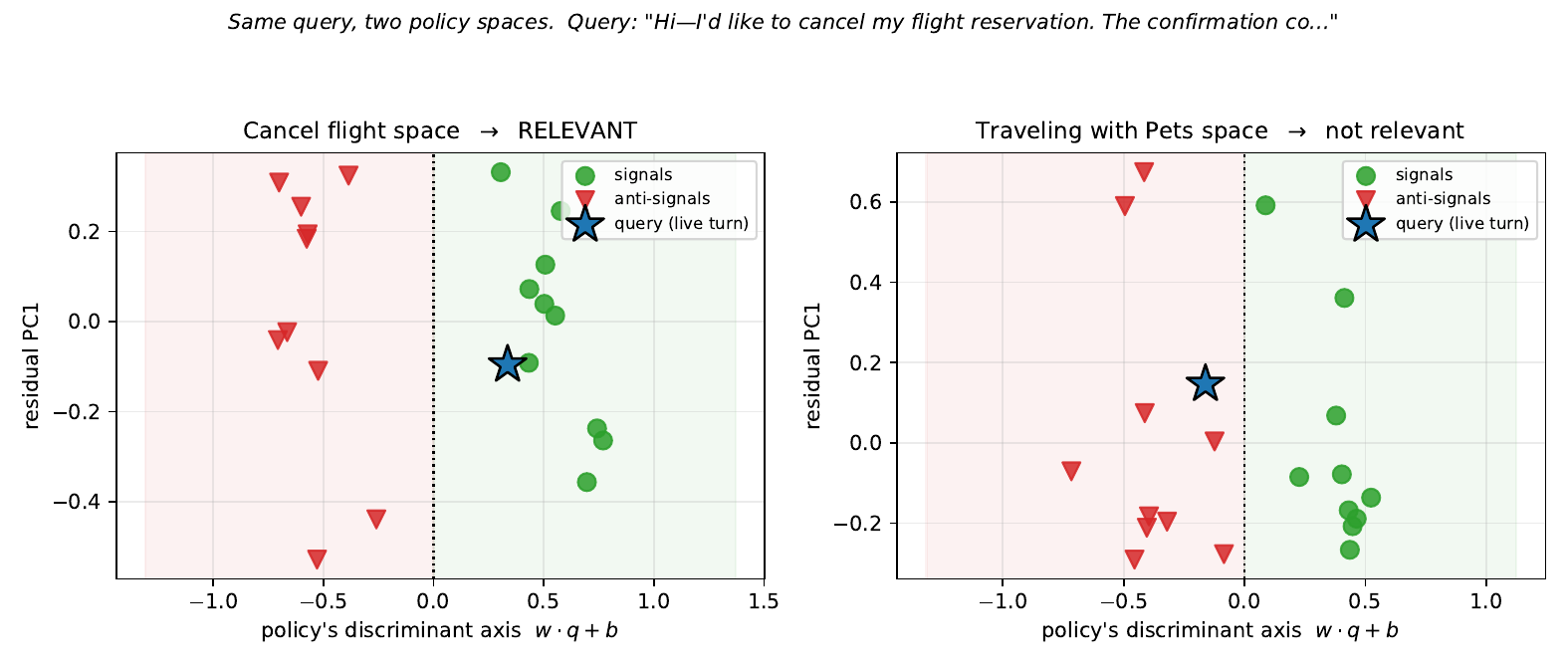}
  \caption{One live conversation turn (\emph{query}, star) projected into two artifacts' classifier spaces ($x$-axis: the artifact's own discriminant $w\!\cdot\!q+b$; $y$-axis: first residual PC of its exemplars). In the \emph{Cancel flight} space the query lands among the signals (relevant); in the \emph{Traveling with Pets} space the same query lands among the anti-signals (not relevant).}
  \label{fig:space}
\end{figure*}

\subsection{Design goals and properties}
\textbf{Recall-first.} As argued above, the loss function of policy discovery is asymmetric. \snq{} therefore targets ${\ge}90\%$ recall per artifact and optimizes precision only subject to that constraint. The retrieved set is a \emph{pre-filter}: the agent's own LLM reads the surviving artifacts and can ignore inapplicable ones cheaply, whereas it can never recover an artifact that was filtered out.

\textbf{Decoupling.} Because artifact $a_j$'s classifier uses only $S_j$ and $\bar S_j$, inventory edits are local: adding, removing, or rewriting one artifact requires generating its exemplars and fitting one classifier (${\sim}2$\,ms), leaving all others untouched. In an earlier variant we trained each artifact against \emph{all other artifacts'} signals, which couples nearby artifacts through their negatives: when two policies sit close in embedding space, each falls in the other's negative set, so the two learned boundaries push their shared region to the negative side of \emph{both}, and a turn genuinely relevant to either is rejected by both, a false negative on each. The effect compounds as the inventory grows and near-duplicates accumulate. Self-contained anti-signals remove the coupling structurally: each artifact's negatives are its own near-misses, so no other policy's presence can suppress it.

\textbf{Near-synonym robustness.} Anti-signals place the decision boundary along an artifact's most confusable neighbors (cancel vs.\ modify vs.\ rebook), where naive cosine retrieval fails.

\textbf{Token efficiency.} All LLM work happens offline ($2N$ generation calls per inventory, re-run only for changed artifacts). Online, a turn costs one embedding call for the new text plus $N$ dot products.

\subsection{Capabilities relative to embedding retrieval}
\label{sec:caps}
Having described the method, we make explicit the structural properties that separate it from ordinary embedding retrieval; \S\ref{sec:results} then quantifies their effect. Table~\ref{tab:caps} compares \snq{} with the two inference-free retrieval families it most directly competes with: \emph{top-$K$ retrieval}, which ranks artifacts by query similarity and returns the $K$ highest, and \emph{cosine-threshold retrieval}, which returns every artifact scoring above a global threshold. Because all three avoid per-turn language-model calls, the decisive question is not cost but capability, namely which properties a method needs in order to serve as a reliable, recall-first policy pre-filter. We identify six.

\begin{table}[t]
\centering\small
\caption{Capabilities of \snq{} relative to the two inference-free embedding-retrieval families it competes with (the baselines of \S\ref{sec:setup}). \fyes: supported; \fno: not supported; \fpart: partial. Each property is defined and motivated in the text; the signal-scored baseline variants (B2, B4) do not change any entry.}
\label{tab:caps}
\begin{tabular}{lccc}
\toprule
Property & top-$K$ & Cosine thr. & \snq{} \\
\midrule
Inference-free per turn        & \fyes  & \fyes & \fyes \\
Adaptive selection size        & \fno   & \fyes & \fyes \\
Uncapped recall                & \fno   & \fyes & \fyes \\
Near-synonym discrimination    & \fno   & \fno  & \fyes \\
Per-artifact decision boundary & \fno   & \fno  & \fyes \\
Portable operating point       & \fpart & \fno  & \fyes \\
\bottomrule
\end{tabular}
\end{table}

\begin{itemize}\itemsep2pt
  \item \emph{Inference-free per turn.} Relevance is decided without any language-model call on the live turn, at the cost of a single embedding lookup. All three methods share this property; it is the common premise that makes the remaining, capability-level distinctions the deciding ones.
  \item \emph{Adaptive selection size.} The count of genuinely applicable policies varies per turn, so a method should return however many clear the relevance bar rather than a fixed number. A ranked cutoff cannot: any constant $K$ is at once too large for simple turns and too small for multi-intent ones, discarding the artifact ranked $K{+}1$ however relevant it is.
  \item \emph{Uncapped recall.} Because the number of relevant policies varies, a fixed cutoff also bounds achievable recall from above (top-$K$ saturates near $66\%$ on our data, \S\ref{sec:results}). A policy pre-filter, whose omissions are correctness failures rather than mere inefficiencies, must be able to approach full recall.
  \item \emph{Near-synonym discrimination.} Policies often share a topical neighborhood (cancellation, modification, rebooking), and a method must separate an artifact from its closest neighbors, where surface similarity peaks. Ranking by cosine similarity, whether against policy text or signals, cannot do this; contrasting signals against hard-negative anti-signals places the boundary there.
  \item \emph{Per-artifact decision boundary.} top-$K$ and cosine thresholding apply one global knob ($K$ or $\tau$) to every artifact, implicitly assuming a shared score scale. \snq{} instead fits an independent boundary per artifact, letting each intercept absorb that artifact's own score offset, which lets a single global margin govern a heterogeneous inventory.
  \item \emph{Portable operating point.} An operating point should transfer across embedding providers and query types without re-tuning. Absolute cosine values reflect each model's geometry, so a threshold must be recalibrated whenever either changes; \snq{}'s margin, measured from each learned boundary, needs only mild per-provider adjustment. top-$K$'s $K$ transfers as an integer but denotes no calibrated relevance level, hence its partial mark.
\end{itemize}

\section{Experimental Setup}
\label{sec:setup}

\textbf{Domain and conversations.} We use the \emph{airline} domain of $\tau^2$-bench~\citep{barres2025tau2}, the leading customer-service agent benchmark suite, maintained by the research department of Sierra.ai~\citep{yao2024taubench}. We take 50 simulated customer conversations produced by a Parlant agent run against the benchmark's user simulator. Every conversation prefix ending at a user turn is a query state: $254$ sub-sequences.

\textbf{Artifact inventory.} $N{=}36$ policies: the 6 sections of the benchmark's airline policy plus 30 additional realistic policies authored for this study (identity verification, baggage claims, unaccompanied minors, fare rules, fraud handling, etc.), including deliberate near-synonym clusters to stress discrimination. Signals and anti-signals ($10{+}10$ per policy) are generated by \texttt{gemini-3.5-flash} from the policy text alone, with no conversation data.

\textbf{Ground truth.} An LLM judge scores every (sub-sequence, policy) pair for relevance on a 1--5 scale: $9{,}144$ labels from \texttt{gemini-3.1-flash-lite}, using the relevance-ranking prompt of Parlant's production engine. Following the engine's own definition, a pair is \emph{relevant} iff score ${\ge}4$. LLM judges are imperfect~\citep{zheng2023judge}; we discuss the implications in \S\ref{sec:discussion}. Metrics are macro-averaged over the 31 policies with nonzero base rate.

\textbf{Baselines.} All methods share the same query views and max-pooling of Eq.~\ref{eq:maxturn} and the same centered-cosine geometry; they differ only in scoring:
\emph{(B1) top-$K$ policy}: $\cos(q, a_j)$ against the policy text, retrieve the top $K$;
\emph{(B2) top-$K$ signal}: $\max_i \cos(q, s_{ji})$, top $K$;
\emph{(B3) threshold policy}: $\cos(q,a_j) > \tau$;
\emph{(B4) threshold signal}: $\max_i \cos(q,s_{ji}) > \tau$.
Sweeping each method's knob traces its recall--precision frontier.

\textbf{Embedders.} The full pipeline is run twice: \texttt{gemini-embed\-ding-001} and OpenAI \texttt{text-embedding-3-large}.

\section{Results}
\label{sec:results}

\begin{figure*}[t]
  \centering
  \includegraphics[width=0.62\textwidth]{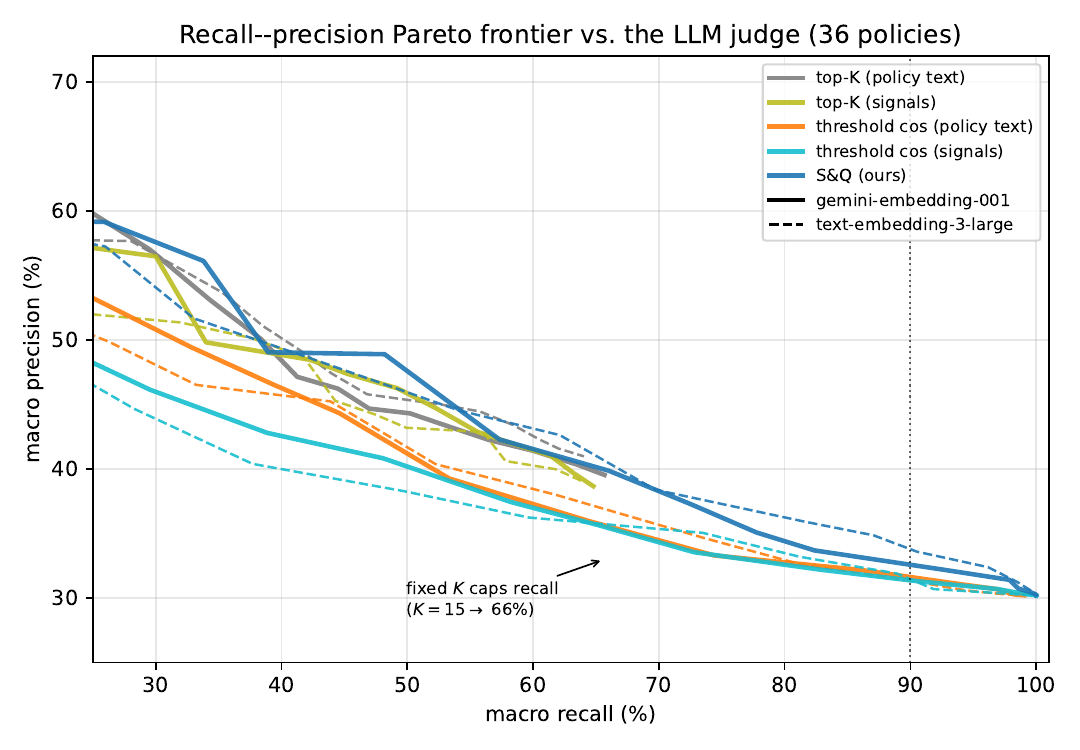}
  \caption{Macro recall--precision Pareto frontiers vs.\ the LLM judge. \snq{} (blue) dominates all four baselines at every recall level, under both embedders (solid: Gemini; dashed: OpenAI). Fixed-$K$ retrieval cannot exceed ${\sim}66\%$ recall because the number of simultaneously relevant policies varies per turn.}
  \label{fig:frontier}
\end{figure*}

\begin{figure*}[t]
  \centering
  \includegraphics[width=0.72\textwidth]{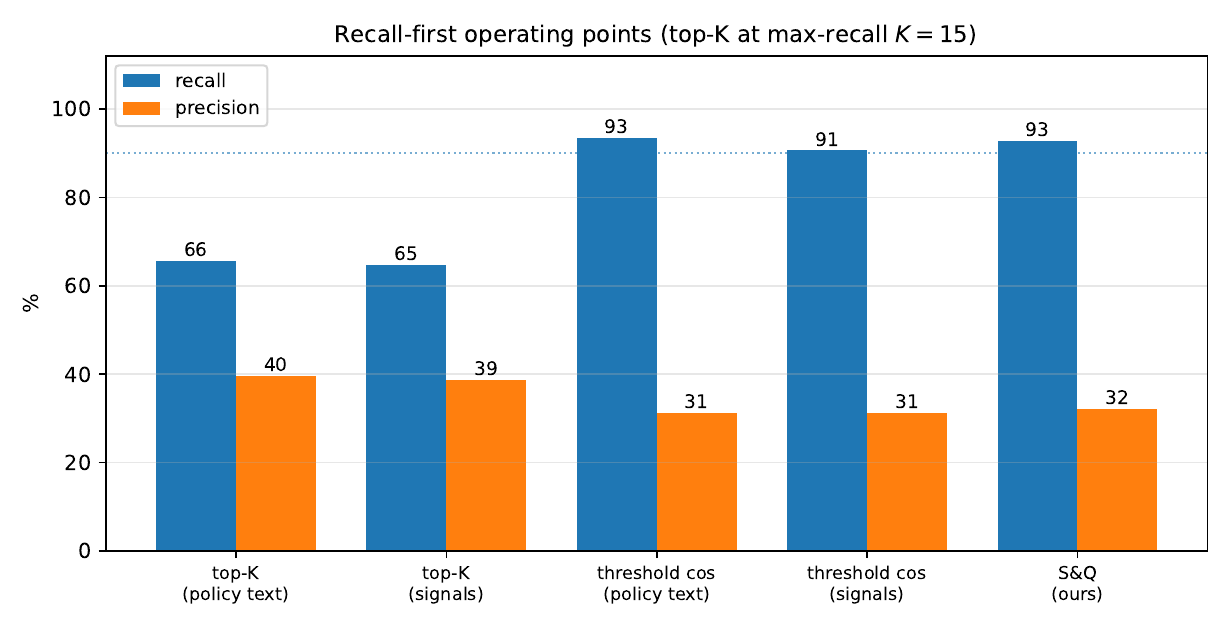}
  \caption{Recall-first operating points (Gemini): best precision subject to macro recall ${\ge}90\%$; top-$K$ methods shown at their maximum-recall setting $K{=}15$, which still fails the recall target.}
  \label{fig:bars}
\end{figure*}

\begin{table}[t]
\centering\small
\caption{Recall-first operating points (best macro precision s.t.\ macro recall ${\ge}90\%$). Top-$K$ methods never reach 90\% recall; shown at $K{=}15$. $^{*}$This operating point is not fixed: \S\ref{sec:scalability} shows \snq{} improves it substantially from exemplars mined from deployment traffic.}
\label{tab:ops}
\begin{tabular}{lcccc}
\toprule
 & \multicolumn{2}{c}{Gemini} & \multicolumn{2}{c}{OpenAI} \\
\cmidrule(lr){2-3}\cmidrule(lr){4-5}
Method & R\,\% & P\,\% & R\,\% & P\,\% \\
\midrule
top-$K$ (policy text) & 66 & 40 & 64 & 41 \\
top-$K$ (signals) & 65 & 39 & 64 & 39 \\
threshold cos (policy) & 93 & 31 & 94 & 31 \\
threshold cos (signals) & 91 & 31 & 92 & 31 \\
\textbf{\snq{} (ours)}$^{*}$ & \textbf{93} & \textbf{32} & \textbf{91} & \textbf{34} \\
\bottomrule
\end{tabular}
\end{table}

\textbf{\snq{} Pareto-dominates the baselines.} Figure~\ref{fig:frontier} shows the frontiers. At every recall level \snq{} yields the highest precision. The margin is widest in the mid-recall range, about ${+}4$ points near 60--70\% recall on both providers, and narrows toward the high-recall extreme, where the two providers diverge: it holds at ${+}2.9$ points on OpenAI but falls to ${+}0.9$ on Gemini. The operating points of Table~\ref{tab:ops} sit at that extreme, which is why they show a clear \snq{} lead on OpenAI ($34\%$ vs.\ $31\%$) but a narrow one on Gemini ($32\%$ vs.\ $31\%$). Unlike top-$K$, whose recall is capped near $66\%$, \snq{} can also be pushed to full recall as $\theta$ decreases.

\textbf{Fixed-$K$ retrieval cannot fit this problem.} Even at $K{=}15$ of 36 policies, top-$K$ recall saturates at ${\sim}66\%$: the number of simultaneously relevant policies varies widely per turn (broad conduct policies are always in play; several topical policies co-activate during multi-intent turns), so no single $K$ fits. Relevance discovery needs per-artifact \emph{decisions}, not a ranked cutoff.

\textbf{The gain comes from the classifier, not the signals.} Threshold retrieval over raw signal embeddings (B4) performs no better than over policy text (B3). The gain appears only when signals are \emph{contrasted} with anti-signals through a trained discriminant, which reorients the decision axis toward what distinguishes the artifact from its neighbors rather than what makes it generically similar to conversation text.

\textbf{Trained boundaries port better than cosine thresholds.} Although threshold-based cosine retrieval reaches decent recall-first operating points (Table~\ref{tab:ops}), the threshold itself is a fragile quantity that is hard to generalize: absolute cosine values are artifacts of each embedding model's geometry and of the query type, not calibrated relevance scores, so a threshold tuned in one configuration does not transfer to another. This is visible even within our benchmark: the best thresholds shift between providers (policy-text $\tau{=}{-}0.025$ on Gemini vs.\ ${-}0.050$ on OpenAI; signals ${+}0.225$ vs.\ ${+}0.275$) and differ substantially between scoring types on the same provider (policy-text vs.\ signal similarities). \snq{} mitigates this: because each artifact's boundary is \emph{learned} from its own exemplars, the intercept $b_j$ absorbs the artifact-, provider-, and query-type-specific score scale, leaving a single global margin $\theta$ whose meaning (distance from each artifact's own decision boundary) is stable. In practice $\theta$ needed only mild per-provider adjustment (${-}0.45$ vs.\ ${-}0.4$), whereas raw-cosine thresholds must be recalibrated from scratch.

\begin{figure*}[!tb]
  \centering
  \includegraphics[width=0.85\textwidth]{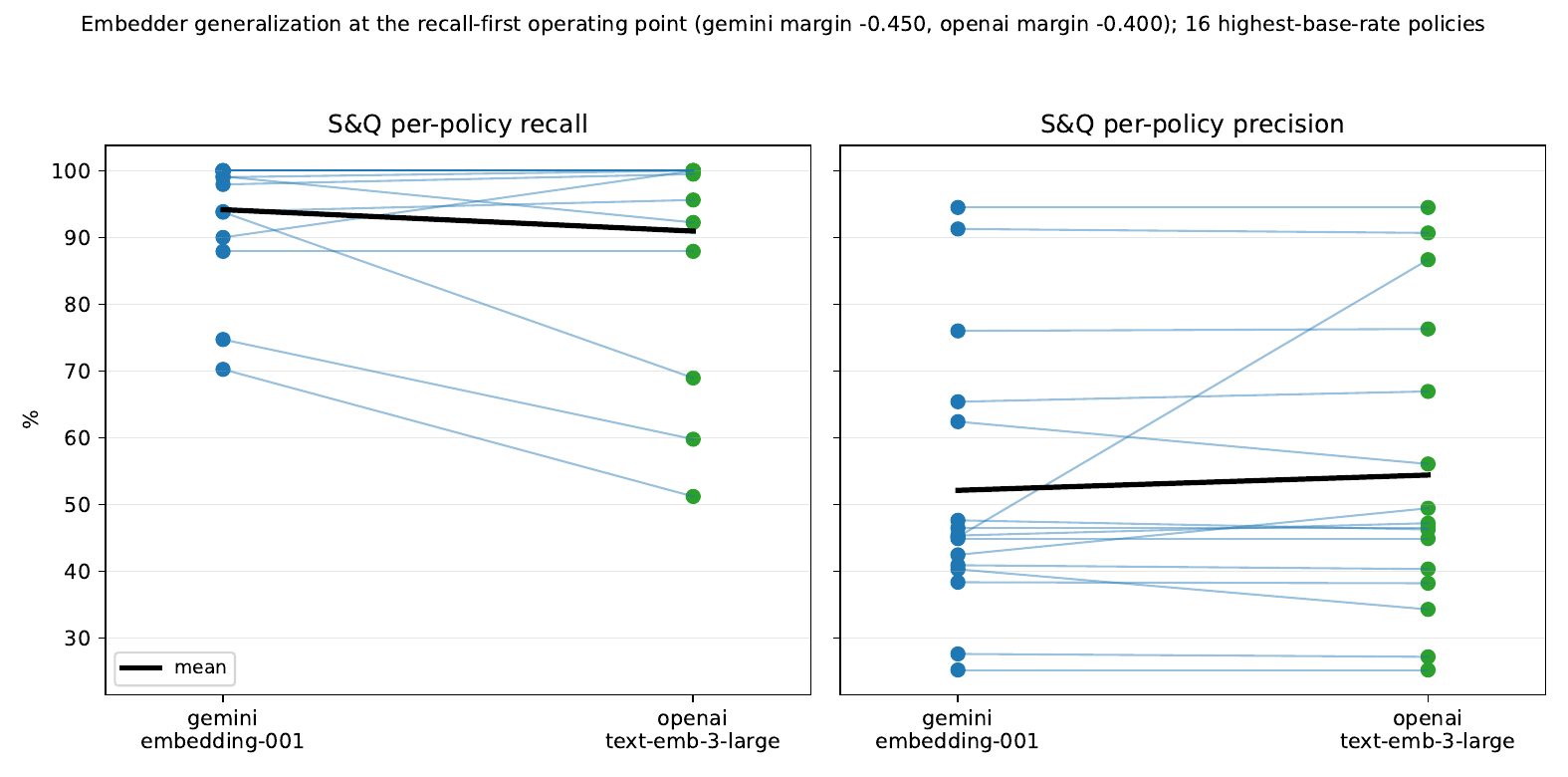}
  \caption{\snq{} per-policy recall (left) and precision (right) under the two embedding providers, at each provider's recall-first operating point (16 highest-base-rate policies; black line: mean).}
  \label{fig:embedders}
\end{figure*}

\begin{figure}[!tb]
  \centering
  \includegraphics[width=\linewidth]{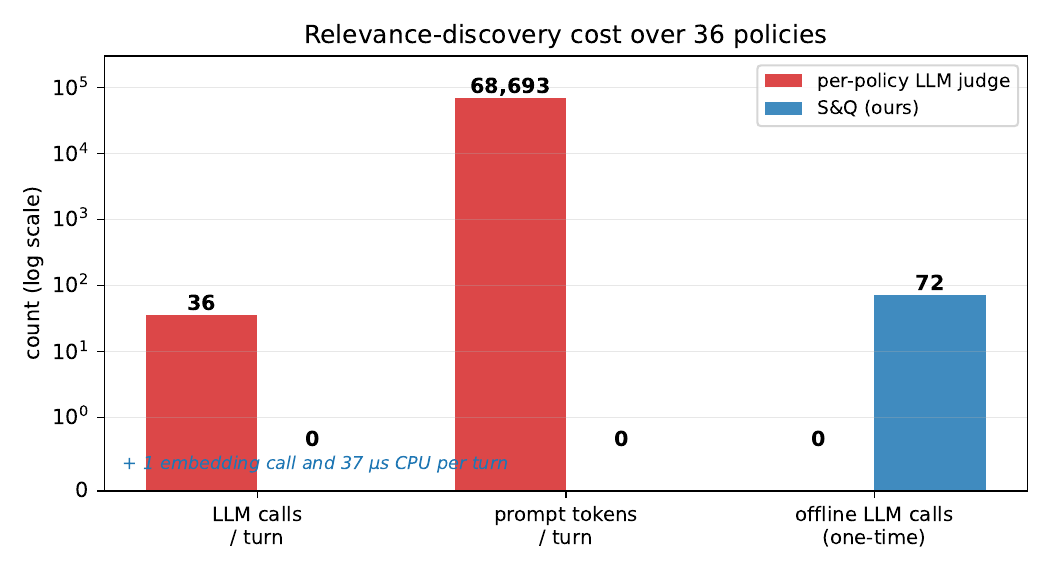}
  \caption{Per-turn relevance-discovery cost (log scale): the per-policy LLM judge vs.\ \snq{}. All of \snq{}'s LLM usage is a one-time offline generation step; its per-turn cost is one embedding call and $37\,\mu$s of CPU.}
  \label{fig:cost}
\end{figure}

\subsection{Generalization across embedders}
Figure~\ref{fig:embedders} shows per-policy \snq{} recall and precision for the 16 highest-base-rate policies under both providers, at each provider's recall-first operating point. The behavior transfers: most policies keep high recall (13 of 16 above $88\%$ on Gemini, 12 on OpenAI), the macro operating point is close across providers (Gemini $93/32$, OpenAI $91/34$), and the method ordering on the frontier is unchanged (Fig.~\ref{fig:frontier}, dashed). Per-policy characteristics do shift with the embedder, a reminder to validate the global margin $\theta$ per provider, but no re-design or re-generation of exemplars was needed.

\subsection{Ablations}
\label{sec:ablation}

\begin{table}[t]
\centering\small
\caption{Ablations: each block varies one component from the full method (\textbf{bold} = default; others held at pooling both, anti-signals, centered, logistic regression, $m{=}10$). AUPRC is the area under the macro recall--precision curve; P@R90 is macro precision at 90\% recall; both are averaged over the 31 occurring policies. G = Gemini, O = OpenAI.}
\label{tab:ablation}
\begin{tabular}{lcccc}
\toprule
 & \multicolumn{2}{c}{AUPRC} & \multicolumn{2}{c}{P@R90 \%} \\
\cmidrule(lr){2-3}\cmidrule(lr){4-5}
Variant & G & O & G & O \\
\midrule
\multicolumn{5}{l}{\emph{Query pooling}}\\
\quad cumulative only          & .526 & .531 & 34 & 34 \\
\quad max-over-turns only      & .462 & .467 & 32 & 33 \\
\quad \textbf{both (maxturn)}  & .471 & .475 & 32 & 34 \\
\addlinespace
\multicolumn{5}{l}{\emph{Negatives}}\\
\quad \textbf{anti-signals}    & .471 & .475 & 32 & 34 \\
\quad generic (other policies) & .501 & .498 & 32 & 33 \\
\quad none (positive-only)     & .423 & .429 & 31 & 31 \\
\addlinespace
\multicolumn{5}{l}{\emph{Centering}}\\
\quad \textbf{on}              & .471 & .475 & 32 & 34 \\
\quad off                      & .463 & .473 & 32 & 33 \\
\addlinespace
\multicolumn{5}{l}{\emph{Classifier}}\\
\quad \textbf{logistic reg.}   & .471 & .475 & 32 & 34 \\
\quad linear SVM               & .463 & .473 & 32 & 33 \\
\quad signed $k$-NN            & .399 & .396 & 30 & 30 \\
\addlinespace
\multicolumn{5}{l}{\emph{Exemplars per policy}}\\
\quad $m{=}5$                  & .436 & .436 & 31 & 33 \\
\quad \textbf{$m{=}10$}        & .471 & .475 & 32 & 34 \\
\bottomrule
\end{tabular}
\end{table}

Table~\ref{tab:ablation} varies one component at a time from the full method, reporting the area under the macro recall--precision curve (AUPRC) and precision at 90\% recall (P@R90) under both embedders. Three findings stand out. \emph{The negatives are what matter.} Removing them, scoring by nearest-signal similarity alone, is the single largest regression (AUPRC $0.47{\to}0.42$; P@R90 $32{\to}31$), confirming that the contrastive discriminant, not the signals in isolation, drives the result (cf.\ baseline B4 in Table~\ref{tab:ops}). Self-contained anti-signals and generic other-policy negatives are near-identical in accuracy (generic is marginally higher), so anti-signals buy their decoupling and near-synonym robustness (\S\ref{sec:method}) at no measurable accuracy cost. \emph{The classifier is not a sensitive choice, for a measurable reason.} Logistic regression and a linear SVM tie (Gemini $0.471$ vs.\ $0.463$; OpenAI $0.475$ vs.\ $0.473$). A linear SVM is the natural challenger here: the exemplar sets are small, high-dimensional, and separable, exactly the regime where its max-margin objective, which ignores easy points far from the boundary, is expected to help. It does not, and a direct measurement shows why. Fit to the same $2m{=}20$ exemplars, the two classifiers learn almost the same direction (mean $\cos(w_{\mathrm{LR}},w_{\mathrm{SVM}}){=}0.98$ under both embedders). The classes are linearly separable for every policy, yet \emph{every} exemplar is a support vector: because anti-signals are deliberately hard near-misses, no example sits outside the margin, so the one situation in which hinge and log loss diverge (their differing treatment of easy, far-from-boundary points) never arises. The hard-negative design that sharpens the boundary also removes the degree of freedom on which SVM could have improved. A signed $k$-NN over the same exemplars, natural given the small per-artifact sample, is clearly worse ($0.40$): a nearest-exemplar rule cannot suppress the pervasive genre similarity between conversations and policy prose that a learned linear boundary discounts. We report logistic regression throughout not for accuracy, which is a tie, but because its probabilistic output gives a deployment more configurable operating knobs: a threshold expressible as a relevance probability, or derived from an explicit false-negative/false-positive cost ratio, where a raw SVM margin offers only an uncalibrated score until it too is passed through a logistic calibration. \emph{Centering and exemplar count are second-order.} Centering gives a small, consistent lift, and doubling the exemplars from $m{=}5$ to $m{=}10$ raises AUPRC monotonically, reinforcing that exemplar generation (\S\ref{sec:discussion}) is the cheapest available lever.

\begin{table}[t]
\centering\small
\caption{Multi-topic stress test: recall (\%) on the recent topic's (B's) policies at the end of a stitched A-then-B conversation, at a matched false-positive rate. Max-over-turns recovers the recent topic that the blended cumulative embedding dilutes; the method's \emph{both} pooling tracks max-over-turns. G = Gemini, O = OpenAI.}
\label{tab:stress}
\begin{tabular}{ccccc}
\toprule
 & \multicolumn{2}{c}{Gemini} & \multicolumn{2}{c}{OpenAI} \\
\cmidrule(lr){2-3}\cmidrule(lr){4-5}
FP rate & cum. & max-turns & cum. & max-turns \\
\midrule
10\% & 34 & 51 & 35 & 57 \\
20\% & 47 & 65 & 47 & 66 \\
30\% & 58 & 70 & 55 & 74 \\
\bottomrule
\end{tabular}
\end{table}

\emph{Pooling addresses a regime the benchmark does not contain.} Max-over-turns does not raise aggregate AUPRC over the cumulative view and here slightly lowers it, but the two guard against different failures. The cumulative view embeds the whole transcript as one vector, blending every topic discussed; when a conversation dwells on one topic and then shifts, the recent topic is a small fraction of that blend, so its policies are diluted, or even pushed toward their own anti-signals by the earlier off-topic content, and a policy relevant only to the latest turns can be missed. Scoring each turn in isolation and taking the max recovers it. We confirm this with a controlled multi-topic stress test: we stitch pairs of different-task conversations into a single A-then-B dialogue and, at B's final turn, measure recall on B's recent-topic policies, taking their ground truth from the ranker's labels for B in isolation. At a matched false-positive rate (each pooling's threshold set to flag the same fraction of irrelevant policies), cumulative pooling recovers $12$ to $19$ fewer points of the recent topic's policies than max-over-turns at a $30\%$ false-positive rate, a gap that holds across false-positive rates from $10$ to $30\%$ (Table~\ref{tab:stress}); and prepending topic~A lowers B's cumulative scores (by ${\sim}0.016$) while leaving the per-turn scores intact. These stitched conversations are synthetic, built to induce the topic shift the primary benchmark lacks, so this is a mechanism demonstration, and the advantage is clearest on B's recent-topic policies in aggregate, weaker on the subset unique to B. The benchmark's own dialogues are short and single-intent (median ${\sim}1{,}600$ characters), so the mechanism seldom fires there and max-over-turns only adds a few false positives; we retain it in the method for multi-topic robustness while noting it is not what earns the benchmark numbers.

\subsection{Comparison with SetFit}
\label{sec:setfit}

\begin{table}[t]
\centering\small
\caption{Fine-tuning the encoder (SetFit) versus \snq{}'s frozen encoder, at the recall-first operating point. On the same local base model (\texttt{all-MiniLM-L6-v2}, 384-dim), fine-tuning does not beat a logistic head on frozen features, and both match the far larger frozen API embedders. Same benchmark, exemplars, and pooling throughout.}
\label{tab:setfit}
\begin{tabular}{llcc}
\toprule
Method & Encoder & R\,\% & P\,\% \\
\midrule
\snq{} (ours)   & frozen Gemini (API)         & 93 & 32 \\
\snq{} (ours)   & frozen OpenAI (API)         & 91 & 34 \\
frozen + logreg & frozen MiniLM (local)       & 91 & 32 \\
SetFit          & fine-tuned MiniLM (local)   & 90 & 31 \\
\bottomrule
\end{tabular}
\end{table}

\snq{} freezes the encoder and fits only a logistic head; a natural question is whether \emph{fine-tuning} the encoder on the same exemplars would do better. SetFit~\citep{tunstall2022setfit} is the direct test: it contrastively fine-tunes a sentence encoder on a few labeled examples before fitting a logistic head. We run it per policy on the same signals and anti-signals, with the identical benchmark and pooling, and, to separate the effect of fine-tuning from the choice of encoder, also run \snq{}'s frozen-encoder recipe on the same local base model. Table~\ref{tab:setfit} shows that fine-tuning does not help: SetFit trails the frozen head on the same encoder ($90/31$ vs.\ $91/32$), and both match the far larger frozen API embedders. With only $2m{=}20$ exemplars per policy, contrastive fine-tuning has little to exploit and no headroom over a linear head on frozen features, consistent with our finding that exemplar coverage, not the classifier or the encoder, is the binding constraint (\S\ref{sec:discussion}). The comparison also reinforces that \snq{} is encoder-agnostic: a small frozen local model suffices, so the method depends on no particular or heavyweight embedder. Fine-tuning would additionally forgo \snq{}'s central efficiency, requiring a trained model per artifact in place of a single embedding call.

\subsection{Cost}
\label{sec:cost}

\begin{table}[t]
\centering\small
\caption{Measured per-turn cost of relevance discovery over $N{=}36$ policies. The LLM-judge row uses the production ranker prompt on a median-length conversation prefix.}
\label{tab:cost}
\begin{tabular}{lcc}
\toprule
 & LLM judge & \snq{} \\
\midrule
LLM calls / turn & 36 & 0 \\
Prompt tokens / turn & ${\sim}69{,}000$ & 0 \\
Embedding calls / turn & 0 & 1 \\
CPU / turn & n/a & $37\,\mu$s \\
\midrule
Offline LLM calls & 0 & $2N{=}72$ (once) \\
Offline training & n/a & 80\,ms (all 36) \\
\bottomrule
\end{tabular}
\end{table}

Table~\ref{tab:cost} and Figure~\ref{fig:cost} quantify the headline claim. Per-policy LLM judging costs ${\sim}69$k prompt tokens and 36 model calls per user turn ($7.6$k characters per policy prompt, measured with the production prompt builder); \snq{} costs one embedding call for the new turn text plus $37\,\mu$s of CPU for all 36 policies, with zero per-turn LLM usage. The entire offline footprint, 72 generation calls and 80\,ms of classifier training, amortizes over every subsequent conversation and is re-incurred only for changed artifacts.

 \snq{} does not fully match the judge's precision at equal recall; it is a pre-filter whose residual false positives are cheap, and which can be composed with a downstream judge over the (much smaller) surviving set when higher precision is required.

\section{Scalability}
\label{sec:scalability}

\begin{figure*}[t]
  \centering
  \includegraphics[width=0.74\textwidth]{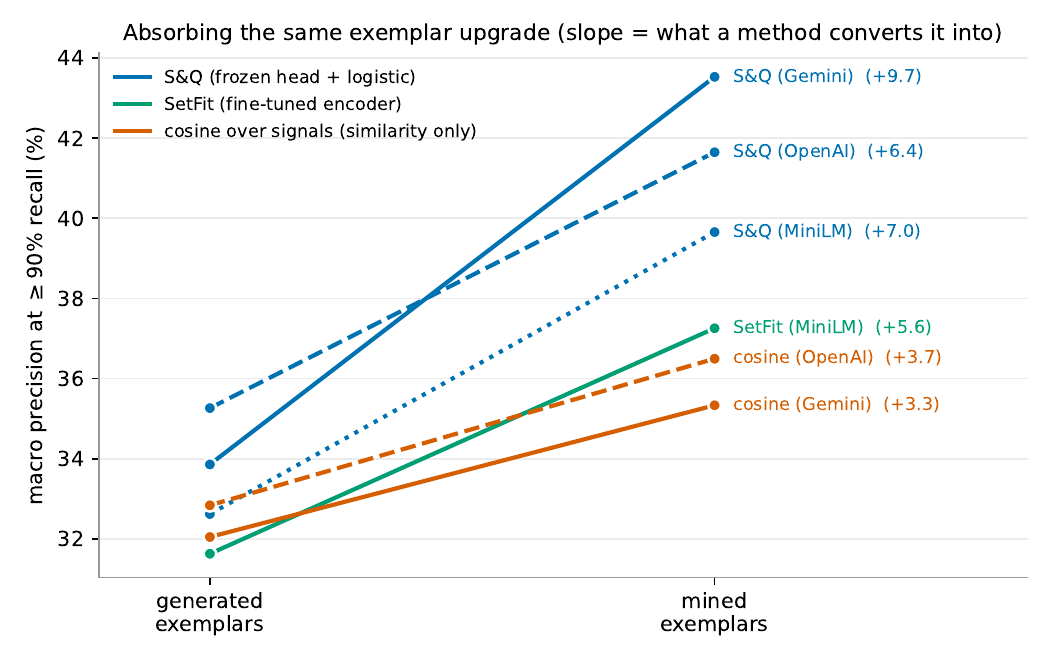}
  \caption{Absorbing the same exemplar upgrade. Every method is fitted first on generated exemplars and then on exemplars mined from held-in conversations, and evaluated on the identical held-out scenarios and artifacts; the \emph{slope} is therefore what each method converts an exemplar improvement into. Colour marks the method family and line style the encoder. Trained discriminants (blue, green) absorb the upgrade; similarity over signals (orange) has no place to put a negative and saturates.}
  \label{fig:absorption}
\end{figure*}

\begin{figure}[t]
  \centering
  \includegraphics[width=\linewidth]{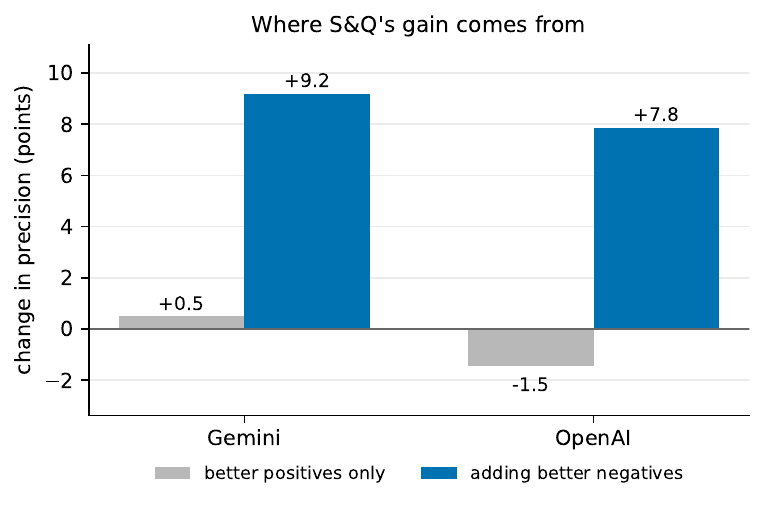}
  \caption{Decomposing \snq{}'s gain. Replacing generated \emph{positives} alone is worth almost nothing; essentially the whole improvement arrives with the mined hard \emph{negatives}, the channel that similarity retrieval cannot express.}
  \label{fig:decomp}
\end{figure}

\begin{figure*}[t]
  \centering
  \includegraphics[width=0.8\textwidth]{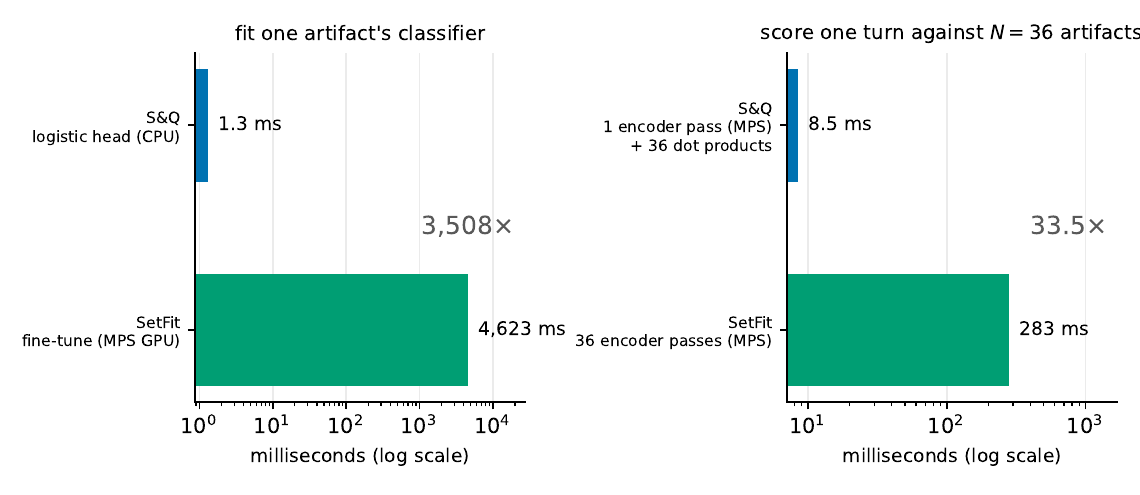}
  \caption{What it costs to own the improvement loop, measured on one machine (Apple M1 Max, 10 cores, macOS) with both methods sharing the same local encoder (\texttt{all-MiniLM-L6-v2}, 384-dim) so the comparison is hardware-matched. SetFit's fine-tuning and \emph{all} encoder forward passes run on the Metal (MPS) GPU backend; \snq{}'s logistic fit runs on CPU in scikit-learn. Left: fitting one artifact's classifier. Right: scoring one live turn against $N{=}36$ artifacts, where \snq{} embeds the turn once and takes $N$ dot products while SetFit must re-embed it once per artifact, since each artifact owns its own fine-tuned encoder. Note the log scale. In the paper's main configuration \snq{} uses a hosted embedding API, so the left-hand bars, not the absolute right-hand latencies, are the deployment-relevant comparison.}
  \label{fig:scalecost}
\end{figure*}

\begin{table*}[t]
\centering\small
\caption{The same exemplar upgrade, absorbed differently. Each method is fitted first on generated exemplars and then on exemplars mined from one half of the scenarios, and evaluated on the held-out other half (\S\ref{sec:scalability}), with artifacts and pooling fixed. The change, not the level, is the quantity of interest: it is what a method converts an exemplar improvement into.}
\label{tab:scalability}
\begin{tabular}{lcc}
\toprule
Method (encoder) & P@R90 \% & AUPRC \\
\midrule
cosine, signals (Gemini) & 32.0 $\to$ 35.3 ({+}3.3) & .459 $\to$ .501 ({+}.042) \\
cosine, signals (OpenAI) & 32.8 $\to$ 36.5 ({+}3.7) & .467 $\to$ .480 ({+}.013) \\
\addlinespace
\snq{} (Gemini)       & 33.9 $\to$ 43.5 (\textbf{{+}9.7}) & .475 $\to$ .602 (\textbf{{+}.127}) \\
\snq{} (OpenAI)       & 35.3 $\to$ 41.6 (\textbf{{+}6.4}) & .496 $\to$ .608 (\textbf{{+}.112}) \\
\snq{}, frozen head (MiniLM) & 32.6 $\to$ 39.7 (\textbf{{+}7.1}) & .457 $\to$ .540 (\textbf{{+}.083}) \\
SetFit, fine-tuned (MiniLM)  & 31.6 $\to$ 37.3 (\textbf{{+}5.7}) & .450 $\to$ .559 (\textbf{{+}.109}) \\
\bottomrule
\end{tabular}
\end{table*}

Every component we varied other than the exemplars proved inert. The classifier family is immaterial: logistic regression and a linear SVM learn nearly the same direction (mean $\cos{=}0.98$) and tie, while a nearest-neighbour rule over the same exemplars is clearly worse (\S\ref{sec:ablation}). The encoder is immaterial: fine-tuning it on the same exemplars does not beat a logistic head on frozen features, and a small local model matches the far larger API embedders (\S\ref{sec:setfit}). Exemplar \emph{count} saturates: extending each artifact's anti-signal pool from ten to thirty, with per-item hardness held constant, leaves precision at $32.3\%$ versus $32.4\%$, because $94\%$ of the additional anti-signals fall within $0.6$ cosine of one already present and the pool's internal diversity does not move. That leaves the exemplars themselves, and specifically their provenance.

We therefore replaced generated exemplars with mined ones. We split the scenarios in half, mined exemplars from one half, and evaluated on the other. Within the mined half a \emph{signal} is the \emph{transition turn}, the user turn at which the judge's score for an artifact first reaches its maximum and therefore the turn that actually triggered it, and an \emph{anti-signal} is a user turn the judge scores as clearly irrelevant; we take four anti-signals per signal, selecting those closest to the artifact's own signals. Everything else, the classifier, the encoder, the pooling, the artifacts, and the held-out items, is unchanged. Table~\ref{tab:scalability} and Fig.~\ref{fig:absorption} give the result: precision at ${\ge}90\%$ recall rises from $33.9\%$ to $43.5\%$ on Gemini and from $35.3\%$ to $41.6\%$ on OpenAI, with AUPRC rising by $0.127$ and $0.112$. The same upgrade on a small local encoder moves precision from $32.6\%$ to $39.7\%$, so the gain belongs to the exemplars rather than to any particular embedding model.

The gain is almost entirely attributable to the \emph{negatives}. Supplying mined positives while retaining generated anti-signals moves precision by ${+}0.5$ points on Gemini and ${-}1.5$ on OpenAI; adding the mined anti-signals on top supplies the remaining ${+}9.2$ and ${+}7.8$ (Fig.~\ref{fig:decomp}). This is the channel that separates \snq{} from similarity retrieval. A score of the form $\max_i \cos(q,s_{ji}) > \tau$ has nowhere to put a near-miss: it can absorb better positives, and doing so does help it, but only to a point (${+}3.3$ and ${+}3.7$ in Table~\ref{tab:scalability}), after which there is nothing further for it to learn. Retrieval against policy text has no exemplar channel at all. Given identical data, \snq{} converts it into roughly three times the precision gain that signal-cosine manages.

This channel is not exclusive to \snq{}. SetFit, which likewise fits a discriminant over positives and negatives, absorbs the same upgrade to a comparable degree (${+}5.7$ points, AUPRC ${+}0.109$), while remaining below a logistic head on frozen features at both exemplar sources (Table~\ref{tab:scalability}). The division is therefore between \emph{trained discriminants} and \emph{similarity scores}, not between \snq{} and everything else. Within its own family \snq{} is no better at absorbing an improvement; it is cheaper at it, which the next paragraph quantifies.

The loop is also cheap and local, which lets it compound. Exemplars of this kind are a by-product of operation: every conversation the agent handles is a candidate. Promoting one costs a single logistic fit of roughly two milliseconds for the affected artifact, and because each artifact's classifier depends only on its own exemplars, nothing else in the inventory is retrained or recalibrated. Per-turn inference cost is unchanged, one embedding call and $N$ dot products, whatever the exemplars contain. Accuracy therefore improves as a function of usage rather than of prompt budget or model size. The contrast with the alternatives is structural. Improving a per-turn LLM ranker means a longer prompt or a larger model, both of which raise the very per-turn cost that motivates this work. Fine-tuning is closer in kind, since it absorbs exemplar improvements nearly as well, but it is considerably more cumbersome to operate, for three reasons. First, per-turn inference stops being constant in the encoder: because each artifact carries its own fine-tuned model, a live turn must be embedded once \emph{per artifact}, $N$ transformer forward passes in place of one shared embedding and $N$ dot products, which forfeits the very property the method exists for. Second, the improvement loop stops being interactive: refitting one artifact after an operator promotes an exemplar takes ${\sim}7$\,ms for a logistic head against ${\sim}5$\,s of gradient fine-tuning on a GPU, some three orders of magnitude (Fig.~\ref{fig:scalecost}), so updates become a batched retraining job rather than an immediate edit. Third, the inventory turns from data into deployed models: $N$ fine-tuned encoders of ${\sim}90$\,MB each to version, ship, and roll back, against ${\sim}12$\,KB per weight vector, produced by a stochastic procedure that need not return the same classifier from the same exemplars, where the logistic fit is convex and has a unique solution. In a governance setting, where an operator must be able to reproduce and explain why an artifact fired, that last difference is not incidental. Fine-tuning further requires a self-hosted, fine-tunable encoder, foreclosing the hosted embedding APIs that \snq{} uses interchangeably.

Three limits qualify the result. Mining requires relevance labels on real conversations, whether from a judge or an operator, so the loop is cheap but not free. Coverage binds before quality does: at a half-and-half mining split only $22$ of the $29$ occurring artifacts had enough mined material to fit on, and the rarest had none, so generated exemplars remain both the bootstrap and the cold-start path. And exemplar genre must be consistent: mixing generated exemplars with mined ones cancels the benefit in \emph{both} directions, which we read as the discriminant learning to separate synthetic text from real conversation instead of separating the artifact from its neighbours. The held-out set here is $131$ sub-sequences, so we report a direction and a magnitude that replicate across two embedding providers rather than precise point estimates.

\section{Discussion}
\label{sec:discussion}

\textbf{How exemplars fail.} Section~\ref{sec:scalability} establishes \emph{that} exemplars are the lever; the failure modes we observed say what makes one set better than another: (i)~\emph{coverage}: 10 anti-signals under-sample the space of irrelevant conversation, so unanticipated off-topic turns can score high (the main precision cost vs.\ many-negative training); (ii)~\emph{hardness}: anti-signals must be near-misses; replacing them with random negatives collapses the discriminant's usefulness on neighbor confusions; (iii)~\emph{domain match}: generated signals are clean single utterances while queries are messy multi-turn conversations, producing the systematic score shift of Fig.~\ref{fig:logreg} that forces $\theta<0$ and per-provider calibration, and which mining exemplars from real conversations largely removes (\S\ref{sec:scalability}). Each of these is a property of the exemplar set, not of the retrieval machinery, which is why we treat signal and anti-signal construction as a first-class problem rather than a preprocessing detail. Beyond the mining loop of \S\ref{sec:scalability}, natural directions include coverage-driven generation (generate until held-out exemplars are classified correctly) and adversarial anti-signal mining against the current classifier.

\textbf{Precision ceilings are real.} Some artifacts are intrinsically faint in embedding space: in our inventory, booking-related turns share vocabulary with the entire flight-administration genre, capping achievable precision at high recall regardless of method; a supervised probe trained directly on labeled conversations confirms the ceiling. A deployment should treat such artifacts with an operator-facing safety valve (e.g.\ pinning must-fire exemplar conversations, then promoting them into the signal set), which we prototype in the underlying framework but do not evaluate here.

\textbf{Deploying \snq{} over long conversations.} Our analysis so far is per turn. A real agentic deployment, though, runs a single conversation across many turns, and the per-turn retrieved sets (\S\ref{sec:method}) accumulate. Because the max-over-turns score of Eq.~\ref{eq:maxturn} is monotone in $t$, an artifact that once crosses $\theta$ keeps a score above it, so the retrieved set only grows. This is the same stickiness that aids multi-topic recall (\S\ref{sec:scalability}), but its cost is that in a long, topic-shifting conversation the injected set drifts, unchecked, back toward the always-in-context regime this paper opens against. The stickiness is thus a property of our design, not only of the LLM judge. A practical deployment must therefore bound the accumulated set, expiring an artifact once enough others have been retrieved and letting it return only on genuinely new evidence rather than on the turns that first activated it. We neither model nor evaluate this here, and our benchmark's short, single-intent conversations do not exercise it. Parlant~4.0 implements a mechanism of this kind.

\textbf{Judge quality and provenance.} Our ground truth is the first-pass relevance ranker of Parlant's engine (\texttt{gemini-3.1-flash-lite}), the same component the deployed engine acts on, not a human panel. It therefore carries the usual LLM-judge caveats~\citep{zheng2023judge} and, importantly, has no measured inter-rater agreement with human evaluators. This is a real limitation: \snq{} is trained to imitate this judge at low cost and latency, so its accuracy is bounded by the judge's, and some of \snq{}'s residual false positives are plausibly judge false negatives. We use the ranker because it is the relevance signal the engine already relies on in practice, and we treat a human-annotated subset as the natural next validation step. We also observed characteristic judge artifacts, a sticky tendency to keep a policy ``relevant'' once its topic appeared and a diffuse ``maybe'' band, which evaluating at the judge's own binary threshold (${\ge}4$) mitigates but does not remove.

\textbf{Limitations.} One domain (airline) and one conversation source ($\tau^2$-bench simulations); 5 of 36 policies never occur in the data and are unevaluated; the recall-first operating point deliberately accepts low precision (${\sim}30\%$), which is appropriate for policy discovery but not for every retrieval use; and the global margin $\theta$, while a single knob, was selected on this benchmark rather than derived from the inventory alone. \snq{} also decides each artifact's relevance independently from the conversation and does not model \emph{inter-artifact dependencies}: a follow-up policy that applies only because another policy fired, and whose own text the current turn does not match, will not be retrieved on that basis. The decoupling that makes the method scale (\S\ref{sec:scalability}) is what precludes this, so dependency handling belongs to an orchestration layer that consumes \snq{}'s output rather than to \snq{} itself.

\subsection{Skill filtering}
\label{sec:skillfilter}
A recall-first pre-filter is most valuable where a false positive is bounded and a false negative is total. Skill-based agents present that asymmetry in unusually clean form. Under the agent-skills standard now common to major agent frameworks~\citep{anthropic2025skills,openai2026codexskills}, a skill is a directory whose entry point carries YAML frontmatter, a name and a description stating when the skill applies, above a Markdown body holding the instructions themselves. Agents load these in tiers: the frontmatter of \emph{every} installed skill is resident from the start, on the order of tens of tokens each, while a skill's body, which may run to thousands of tokens, is read only once the agent decides to invoke it.

That tiering makes the setting a natural fit. As a library grows to hundreds of skills, the resident frontmatter becomes a standing context cost in its own right, and the agent must choose among many superficially similar descriptions, which is precisely the regime in which instruction adherence degrades~\citep{liu2024lost,harada2025curse}. \snq{} applies to the frontmatter tier without modification: the artifact text is the skill's name and description, its signals are user messages that should bring the skill into consideration, and its anti-signals are the near-misses that belong to a neighbouring skill. The filter decides which frontmatter entries enter the prompt at all.

Recall-first is the right operating point here because \snq{} is not deciding to \emph{run} anything. A false positive costs one frontmatter entry, tens of tokens, which the agent dismisses after reading a sentence; it does not pull in the skill's body, where the real context cost lives. A false negative is unrecoverable: a skill whose description never reaches the prompt cannot be invoked, however relevant it was. Precision is thereby delegated to the stage that can afford it, the agent's own reading of the surviving descriptions, while \snq{} supplies the one property that stage cannot recover on its own, namely that the right candidate is present at all.

Existing practice retrieves skills by similarity against their descriptions: Voyager, for instance, keys a skill library by the embedding of each skill's description and retrieves by nearest neighbour~\citep{wang2023voyager}. That is the positives-only regime of our B2 and B4 baselines, and our results bear on it directly, since similarity against descriptions cannot represent the near-miss that separates a skill from its closest neighbour, and absorbs exemplar improvements at roughly a third the rate of a discriminant (\S\ref{sec:scalability}). We have not evaluated \snq{} on a skill corpus; we present this as an application whose shape the method fits, not as a measured result.

\section{Related Work}

\textbf{Per-turn context selection.} The problem \snq{} addresses, deciding which of many candidate instructions or artifacts to place in the prompt at each turn, recurs across agent settings under different names. Two strategies dominate. Embedding \emph{similarity} against artifact descriptions is cheap and LLM-free, and underlies tool and skill retrieval~\citep{patil2023gorilla,qin2023toolllm}, dynamic tool management across a conversation~\citep{lumer2025memtool}, in-context example selection~\citep{wang2025dice}, and long-horizon memory~\citep{packer2023memgpt}. It is exactly the regime of our B2 and B4 baselines, and our results show where it fails: on the near-misses a similarity score cannot separate from true matches. LLM \emph{ranking}, judging each candidate per turn, is more accurate but pays the per-turn token and latency cost that motivates this work (\S\ref{sec:cost}), and is the ranker we take as ground truth. A complementary line \emph{compresses} a fixed context rather than selecting its contents, dropping low-information tokens to shorten the prompt~\citep{jiang2023llmlingua}. Compression is orthogonal to \snq{} and, for normative policies, risky, since a partially dropped rule can invert the behaviour it was meant to govern, whereas \snq{} injects each selected artifact intact. \snq{} sits at the cheap, LLM-free end of this space but replaces similarity with a recall-first, per-artifact discriminant, buying the near-miss discrimination that plain retrieval lacks.

\textbf{Retrieval machinery.} Dense retrieval represents queries and documents in a shared embedding space~\citep{karpukhin2020dpr,reimers2019sbert}, typically for retrieval-augmented generation~\citep{lewis2020rag}. \snq{} differs in objective (per-artifact binary \emph{decisions} under a recall constraint, not top-$K$ ranking) and in bridging the query--document genre gap from the document side: signals are LLM-predicted activating queries, the exemplar-level analogue of Doc2Query~\citep{nogueira2019doc2query} and of synthetic-query domain adaptation~\citep{wang2022gpl}, while HyDE~\citep{gao2023hyde} attacks the same gap from the query side with hypothetical documents. Unlike these, \snq{} additionally generates \emph{hard negative} exemplars and trains a discriminant per artifact, bringing contrastive supervision~\citep{gao2021simcse} to inference-free per-artifact classification. The leverage of hard negatives is well established on the retrieval side: the passages most likely to distract the model reading them are the topically adjacent ones, and deliberately mining them improves robustness~\citep{amiraz2025distracting}, a principle anti-signals carry to artifact selection by placing each classifier's boundary along its most confusable neighbors. Embedding anisotropy and its correction motivate our centering step~\citep{ethayarajh2019contextual,mu2018allbutthetop,conneau2018word}. We evaluate on the conversational agent benchmarks $\tau$-bench and $\tau^2$-bench~\citep{yao2024taubench,barres2025tau2}.

\section{Conclusion}
Signals and Queries reframes agentic context discovery as a set of tiny, decoupled, recall-first classifiers over LLM-generated exemplars: all the LLM work is spent offline where it amortizes, and per-turn discovery becomes an embedding lookup. On a 36-policy benchmark judged by the production relevance ranker it dominates embedding-retrieval baselines at every recall level, across two embedding providers, at ${\sim}10^{5}\times$ lower per-turn cost than per-policy LLM judging. The method ships as part of Parlant~4.0~\citep{parlant}. Its clearest path forward is better exemplars: generation strategies that maximize coverage and hardness, and operator feedback loops that promote real missed conversations into the signal set.

\section{Algorithm Summary}
\label{sec:pseudocode}

We give pseudocode for the two phases of \snq{}. Offline construction (Alg.~\ref{alg:offline}) runs once per inventory and is re-run only for changed artifacts. Online discovery (Alg.~\ref{alg:online}) runs on each user turn and issues no LLM calls. Notation follows \S\ref{sec:method}: $E$ is the embedding model, $G$ an LLM used only offline, $\hat{v}=v/\lVert v\rVert$, and $c$ the inventory centroid.

\begin{algorithm}[H]
\small
\caption{Offline construction (per inventory)}
\label{alg:offline}
\begin{algorithmic}[1]
\Require artifacts $\mathcal{A}=\{a_1,\dots,a_N\}$; embedder $E$; generator $G$
\For{each artifact $a_j \in \mathcal{A}$}
  \State $S_j \gets G.\textsc{GenSignals}(a_j)$ \Comment{$m$ activating msgs}
  \State $\bar S_j \gets G.\textsc{GenAntiSignals}(a_j)$ \Comment{$m$ near-misses}
\EndFor
\State $c \gets \frac{1}{N}\sum_{j} E(\textsc{text}(a_j))$ \Comment{centroid}
\For{each artifact $a_j \in \mathcal{A}$}
  \State $P \gets \{\,\widehat{E(s)-c} : s \in S_j\,\}$ \Comment{positives}
  \State $Q \gets \{\,\widehat{E(s)-c} : s \in \bar S_j\,\}$ \Comment{negatives}
  \State $(w_j,b_j) \gets \textsc{FitLogReg}(P{:}1,\ Q{:}0)$
\EndFor
\State \Return classifiers $\{(w_j,b_j)\}_{j=1}^N$, centroid $c$
\end{algorithmic}
\end{algorithm}

\begin{algorithm}[H]
\small
\caption{Online discovery (per user turn $t$)}
\label{alg:online}
\begin{algorithmic}[1]
\Require turns $u_1{,}\dots{,}u_t$; classifiers $\{(w_j,b_j)\}$; $c$; margin $\theta$
\State $q_{\mathrm{cum}} \gets \widehat{E(\textsc{render}(u_{1:t})) - c}$ \Comment{cumulative}
\State $q_i \gets \widehat{E(u_i) - c}$ \textbf{ for } $i=1,\dots,t$ \Comment{per-turn}
\State $\mathcal{R} \gets \varnothing$
\For{each artifact $a_j$}
  \State $z_j \gets \max\!\big(w_j\!\cdot\!q_{\mathrm{cum}},\ \max_{i\le t} w_j\!\cdot\!q_i\big) + b_j$
  \Statex \hfill\Comment{max over turns}
  \If{$z_j > \theta$} \State $\mathcal{R} \gets \mathcal{R} \cup \{a_j\}$ \EndIf
\EndFor
\State \Return $\mathcal{R}$ \Comment{artifacts to inject into context}
\end{algorithmic}
\end{algorithm}

Only $q_{\mathrm{cum}}$ and the newest $q_t$ are embedded per turn (earlier $q_i$ are cached from prior turns), so a turn costs one embedding call plus $N$ dot products. The recall-first regime uses $\theta<0$; per-provider values are given in \S\ref{sec:results}. The operator safety valve of \S\ref{sec:discussion} augments line~6 of Alg.~\ref{alg:online} with a disjunctive test against pinned exemplar conversations, and promotes recurring pins into $S_j$ (Alg.~\ref{alg:offline}, line~2).

\section*{Acknowledgements}
We thank Alex Shtoff \textless\texttt{alexander.shtoff@tii.ae}\textgreater{} for a careful review of an earlier draft. His feedback motivated the ablation study of \S\ref{sec:ablation}: in particular, the comparison of the logistic classifier against a linear SVM and a nearest-neighbor variant, and the accompanying analysis of why the classifier choice is immaterial in this small, separable, hard-negative regime. His comments also sharpened the paper's empirical framing around baselines and ablations. We thank Vishal Ahuja \textless\texttt{vishal.x.ahuja@chase.com}\textgreater{} for suggesting the comparison against SetFit~\citep{tunstall2022setfit} in \S\ref{sec:setfit}. We thank Sarthak Dalabehera \textless\texttt{s.dalabehera@rippling.com}\textgreater{} for the skill-filtering application of \S\ref{sec:skillfilter}. We thank Bar Karov \textless\texttt{barkrv@gmail.com}\textgreater{} for a careful review that reconciled the frontier plot with the operating-point table (\S\ref{sec:results}) and prompted the treatment of related work on per-turn context selection. We thank Shailen Kumar Dalbehera \textless\texttt{ShailenKumar.Dalbehera@evernorth.com}\textgreater{} for raising the provenance and human validation of the LLM judge and the accumulation of retrieved policies over long conversations, which motivated the discussion of pruning (\S\ref{sec:discussion}).

\bibliographystyle{unsrtnat}
\bibliography{refs}

\end{document}